\documentclass[acmtog, authorversion = false]{acmart}
\renewcommand\footnotetextcopyrightpermission[1]{} 
\usepackage{booktabs} 

\setcitestyle{square}
\usepackage[noend]{algpseudocode}
\usepackage{algorithm}
\usepackage{amsmath}
\usepackage{amsfonts,amsthm}
\usepackage{bm}

\makeatletter
\algrenewcommand\ALG@beginalgorithmic{\footnotesize}
\makeatother

\newcommand{\beq}{\begin{equation}}
\newcommand{\eeq}{\end{equation}}

\newcommand{\bea}{\begin{eqnarray}}
\newcommand{\eea}{\end{eqnarray}}

\newcommand*\diff{\mathop{}\!\mathrm{d}}

\pagecolor{white}

\begin{document}

\title{\LARGE \bf
Mid-flight Forecasting for CPA Lines in Online Advertising \\
\small 41st International Symposium on Forecasting,  June 27-30, 2021}

\author{Hao He}
\affiliation{%
 \institution{Demand Platforms R\&D, Verizon Media}
 \streetaddress{701 First Avenue}
 \city{Sunnyvale}
 \postcode{94089}
 \state{CA}
}
\email{hao.he@verizonmedia.com}

\author{Tian Zhou}
\affiliation{%
 \institution{Demand Platforms R\&D, Verizon Media}
 \streetaddress{701 First Avenue}
 \city{Sunnyvale}
 \postcode{94089}
 \state{CA}
}
\email{tian.zhou@verizonmedia.com}

\author{Lihua Ren}
\affiliation{%
 \institution{Demand Platforms R\&D, Verizon Media}
 \streetaddress{701 First Avenue}
 \city{Sunnyvale}
 \postcode{94089}
 \state{CA}
}
\email{lihua.ren@verizonmedia.com}

\author{Niklas Karlsson}
\affiliation{%
 \institution{Demand Platforms R\&D, Verizon Media}
 \streetaddress{701 First Avenue}
 \city{Sunnyvale}
 \postcode{94089}
 \state{CA}
}
\email{niklas.karlsson@verizonmedia.com}

\author{Aaron Flores}
\affiliation{%
 \institution{Demand Platforms R\&D, Verizon Media}
 \streetaddress{701 First Avenue}
 \city{Sunnyvale}
 \postcode{94089}
 \state{CA}
}
\email{aaron.flores@verizonmedia.com}

\begin{abstract}
For Verizon Media \textit{Demand Side Platform} (DSP), forecasting of ad campaign performance not only feeds key information to the optimization server to allow the system to operate on a high performance mode, but also produces actionable insights to the advertisers. In this paper,  the forecasting problem for CPA lines in the middle of the flight is investigated by taking the bidding mechanism into account. The proposed methodology generates relationships between various key performance metrics and optimization signal. It can also be used to estimate the sensitivity of ad campaign performance metrics to the adjustments of optimization signal, which is important to the design of campaign management system. The relationship between advertiser spend and \textit{effective Cost Per Action} (eCPA) is also characterized, which serves as a guidance for mid-flight line adjustment to the advertisers. Several practical issues in implementation, such as downsampling of the dataset, are also discussed in the paper.  At last, the forecasting results are validated against actual deliveries and demonstrates promising accuracy.

\end{abstract}


\thanks{

Author's affiliation: Demand Platform R\&D, Verizon Media.}

\maketitle

\section{Introduction} \label{sec:intro}

The campaign management system of Verizon Media DSP implements a feedback-based control system \cite{kandzh13}, which adjusts the bid prices used to bid for impressions.
Different aspects of how the optimization problem is formulated and solved as a control problem  can be found in ~\cite{karlsson2016control}, \cite{guo2017model}, \cite{sang2018feedback}, \cite{karlsson2016systems}. The fact that the plant \footnote{Plant is the term used to indicate the relation between an input signal and the output signal of a system without feedback, determined by the properties of the whole online advertising ecosystem }
is unknown, dynamic, nonlinear, and in general discontinuous is a characteristic property of online advertising processes and is a fundamental challenge in the development of feedback control solutions. In order to design feedback control algorithms for the optimization system, it is valuable to have a model of the input-output relationship, which is referred to as the control response curves in this paper. The key performance metrics for CPA lines include impressions, total advertiser spend, eCPA, etc. Besides,  the problem of estimating the sensitivity of the output (mainly the total advertiser spend) to the control signal, which usually termed as the plant gain in control community, is also crucial to controller design and challenging due to the discontinuity of the plant. 

Different approaches at indirectly estimating and controlling the plant are proposed in \cite{karlsson2014adaptive}, \cite{karlsson2017plant}, \cite{guo2017model}. The former two of these papers make use of bid randomization, but all are based exclusively on local feedback information to estimate the plant gain. A systematic methodology for offline modeling of advertising plants is proposed in \cite{mardanlou2017statistical}. The methodology is conveniently used to simulate realistic plants in a test bed for control algorithms. In this paper, we propose an analytical solution of advertiser spend as a function of control signal by modeling the event rate of available impressions using a \textit{Gaussian Mixture Model} (GMM). In this way, the plant gain can be directly derived based on its definition.

To the advertisers, it is always of interest to know how different performance metrics interact with each other. For example, how does the eCPA change as the budget increase? How much incremental budget advertisers can spend without violating their eCPA constraint? The above questions can be answered by examining different control response curves, since they are all monotonically increasing functions of the control signal. Building the different control response curves will provide valuable and actionable information to the advertisers and help them managing their campaigns more efficiently and gradually draw more budget to Verizon Media DSP.

The rest of this paper is organized as follows. The problem of forecasting for CPA lines in the middle of the flight is formulated mathematically in Section \ref{sec:probf}. The detailed steps of the algorithm is described in Section \ref{sec:approach}. Experimental results are provided in Section \ref{sec:results} and implementation of the algorithm in production is provided in Section \ref{sec:imp}. At last, in Section \ref{sec:conclusion}, we conclude and propose future work.

\section{Problem Formulation}\label{sec:probf}

In Verizon Media DSP, there are two auction stages, one is internal auction within the DSP and the other is external auction that happens in ad exchanges. In internal auction stage, multiple lines bid on the same impression by submitting a bid price, called score. The highest bidder wins the chance to submit a bid to the ad exchange. In the second stage, the winner submits a bid price to external ad exchanges.
We assume that the auction is \emph{second-price}~\cite{AuctionTheoryKrishna2002} cost model, which means that the winning bidder pays an amount that equals to the highest competing bid for the impression. In practice, partial of the traffic is under a \textit{first-price} model where the winning bidder pays an amount that equals to its bid price.  But for traffic under first price model, the DSP try to mimic the behavior of second price auctions by implementing a scheme called bid shading \cite{karlsson2019wine}, which tries to reduce the bid as close to the highest competing bid as possible so that the winner bidder will still be able to win but only pay an amount that almost equals the highest competing price. In this work, we assume perfect shading so that all traffic can be viewed as under second price model. The bid price submitted to both internal and external auctions are equal, and is denoted by $b$.

Suppose there is a set of impressions, denoted by $\Omega:=\{1, 2, \dots,i,\dots, N\}$, that meet the targeting of a line, we call the impressions in $\Omega$ \textit{available impression} for the line. For every impression $i$, we use $b^*_i$ to denote the \textit{highest competing price}. The \textit{event rate}, defined as the probability of an awarded impression converting to a conversion, is denoted by $e_i$. The \textit{advertiser cost}, defined as the amount that advertiser pays to impression $i$, is denoted by $b^c_i$. We assume that for available impression of the line, the three variables are independently draw from the same underlying joint distribution, namely $p(b^*_i, e_i, b^c_i) = p(b^*, e, b^c), \forall i \in \Omega$ where $p(\cdot)$ is the \textit{Probability Density Function} (PDF). 

%

It is well known \cite{karlsson2016control} that a value-maximizing bidding strategy is to bid proportional to the value of the impression, which is the even rate $e_i$. We define a \textit{control based bid price}, represented by $b^u_i$, as follows 
\beq \label{eq:controlPrice}
b_i^u = uge_i
\eeq
where $u$ is the line level \textit{control signal}, $g$ is the \textit{goal amount} defined by the advertiser, representing the maximum cost of a conversion. 
The \textit{bid price} $b_i$ for impression $i$ is calculated based on $b^u_i$, with some line level deterministic transformation as will be described in details in Section \ref{sec:bidAdj}

For CPA lines, the eCPA, denoted by $v$, is an important measurement of performance and is defined as the average cost per conversion.
Based on the above model, the number of impressions, total advertiser spend, number of conversions and eCPA can be expressed as the following 
\bea
\label{eq:kpi}
n_I &=& N\mbox{E}_{e, b^*} \left[ \mathbb{I}_{\{b >b^* \} } \right]  \nonumber\\
c &=&  N \mbox{E}_{e, b^*, b^{c}} \left[b^{c} \mathbb{I}_{\{b >b^* \} } \right]  \nonumber\\
n_A & =& N\mbox{E}_{e, b^*, b^{c}} \left[e \mathbb{I}_{\{b >b^* \} } \right] \nonumber\\
v & =& \frac{c}{n_A}
\eea
where $\mathbb{I}_{\{\cdot\}}$ is the indicator function, $\mbox{E}_{X_1, X_2, \dots}[f(x_1, x_2, \dots)]$ represents the expectation of $f(x_1, x_2, \dots)$ over the distribution of $X_1, X_2, \dots$.
The focus of this paper is to forecast the relationship between control signal $u$ with key performance metrics, defined in Eq. (\ref{eq:kpi}). 

\subsection{Dataset} \label{sec:dataset}
The datasets that we use are internal auction data \footnote{which is located in \path{/projects/kite/prod/internal/core/raw_internal_auction/5m} on db} along with some other line metadata from kite database. The internal auction data is 1:4 sampled from raw bid requests. For each bid request, it records the lines whose targeting criteria is met along with the predicted event rate, internal auction bid price by that line. Thus, for each line, we have a 1:4 sampled pool of its available impressions, denoted by $\Omega' := [1, 2, \dots, N']$. 
And for each available impression, we have the four tuples $[e_i, b^{s}_i, b^*_i, b^{c}_i]$ in which $b^s_i$ represents \textit{internal auction bid price} for impression $i$ from the line. 

Note that the highest competing bid $ b^*_i$ is derived. If the line won, the $b^*_i$ is the maximum of second highest bid price in internal auction and the inventory cost for impression $i$; otherwise, $b^*_i$ is the highest internal auction bid price. 

It also needs to be noted that, a \textit{pacing signal}, representing the probability that a line responses to each impression and denoted by $a \in[0,1]$, is applied in the production, meaning that even an impression is considered to be available, with probability $a$ it appears in $\Omega'$.
Thus, to compensate for any $a < 1$, we proposed the normalization step in Section \ref{sec:normalization}. 
\section{Methodology} \label{sec:approach}
\subsection{Bid Adjustment}\label{sec:bidAdj}
Usually, advertisers will set a \textit{max bid} representing the maximum value of bid price that they would pay to any impressions, denoted by $b_{max}$. The control based bid price in Eq. (\ref{eq:controlPrice}) is capped by the max bid. 
The bid price $b_i$ is after further deducting fees and bid shaving from the capped value. We use $\theta_1 \in [0,1]$ to represent the multiplicative coefficient and use $\theta_0 \in [0,1]$ to represent the additive coefficient. Thus, throughout the paper we use the following to represent the overall bidding strategy in production
\bea
\label{eq:linearModel}
b_i &=& f(e_i, u|\theta_1, \theta_0) = \theta_1 \min{ \{b_i^{u}, b_{max} \}} - \theta_0 \nonumber\\
&=& \theta_1\min{\{ uge_i, b_{max} \} } - \theta_0
\eea
The line level parameters $\bm{\theta} := [\theta_1, \theta_0]$ need to learn from historical data \footnote{The parameters  can also be queried from databases, but it adds challenge to building data pipeline. Thus, we choose to estimate directly from existing dataset which turns out to be more efficient.}. The linear model parameterized by $\bm{\theta}$ in Eq. (\ref{eq:linearModel}) is to describe the procedures of deducting fees and bid shaving.

%

The parameters are learnt from the historical data pair event rate $e_i$ and internal auction score $b_i^{s}$ for all $i \in \Omega'$. The control signal in Eq. (\ref{eq:linearModel}) corresponds to the daily average control value when the training set $\Omega'$ is collected. The parameter $\bm{\theta}$ is obtained by solving the following constrained minimization problem

\begin{equation*}
\begin{aligned}
& \underset{\bm{\theta}}{\text{minimize}}
& &\frac{1}{2} \sum_{i=1}^{N'} \| f(e_i, u|\bm{\theta}) -b^s_i \|^2 \\
& \text{subject to}
& & \bm{\theta} \geq 0, \bm{\theta} \leq 1.\\
\end{aligned}
\end{equation*}
The constraints are interpreted from reasonable fees-deducting methods used in the DSP and are added to the optimization problem to avoid extremely abnormal estimates due to noise training dataset. Having an estimated $\hat{\bm{\theta}}$, we are able to apply Eq. (\ref{eq:linearModel}) to calculate bid price based on the event rate.


\subsection{Event Rate Modeling}
\label{sec:eModel}

\begin{figure}
\centering
\includegraphics[width=.7\columnwidth,height=!]{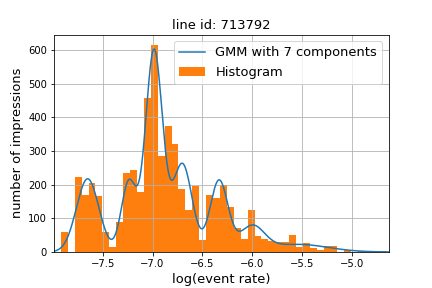}
\caption{Histogram of event rate and the fitted GMM}
\label{fig:gmm}
\end{figure}

Through extensive experiments, we found that the correlationship (Pearson's) between the $e$ and $b^*$ is very slight, with a median of $0.088$ for all 9968 active CPA lines in a single day. Similarly for the correlationship between $e$ and $b^c$.
Thus, we model the distribution of event rate independently from $b^*$ and $b^c$ ($b^*$ and $b^c$ are highly correlated based on the cost model). We choose GMM as a parametric model by observing the clustering behavior of data shown in Fig. \ref{fig:gmm}. The model is explicitly expressed as the following 
\bea
p(e|\bm{\theta}_e) = \sum_{k = 1}^K \pi_k \mathcal{N}(e|\mu_k, \sigma_k)
\eea
where $\pi_k, \mu_k, \sigma_k$ respectively represent the weight, mean and covariance of $k$th Gaussian component and  $K$ is the number of Gaussian components in the model. The parameters $\bm{\theta}_e: =[\pi_k, \mu_k, \sigma_k | k = 1, \dots, K ]$ are estimated using \textit{Expectation Maximization} (EM) algorithm, a well known algorithm for GMM parameter estimation. To choose the proper value of  the hyper parameter $K$, we use \textit{Bayesian Information Criterion} (BIC), which avoids overfitting by adding a penalty term for the number of parameters in the model. 

\subsection{Forecasting}
\subsubsection{Normalization} \label{sec:normalization}
One important assumption that we made in this work is that the number of available impressions for a line is static over time given fixed line targeting, which is quite reasonable, especially considering the huge amount traffic in the DSP.  Thus, we take the number of available impressions of today as the forecast for tomorrow. Let's denote the 5-min available impressions in a day by $\bm{N}':=[N'_1, \dots, N'_{288}]$, and denote the 5-min pacing signal by $\bm{a}:=[a_1, \dots, a_{288}]$. We first adjust the available impressions at non-zero pacing signal to obtain a $N'_j$ corresponding to $a_j = 1$. To facilitate the explanation, we define a set $\Omega_a$ representing the index of all non-zero pacing rate, i.e., $\Omega_a:= [j| a_j > 0]$. 
\bea
N_j = \frac{N'_j}{\sqrt{a_j}}, \forall j \in \Omega_a
\eea
Then, we compensate for the \textit{Time of Day} (TOD) pattern for where the pacing signal is zero due to either day parting, hard stop cap or any other reasons. The TOD pattern of Verizon Media DSP is modeled by a 2-term truncated Fourier series as follows \cite{hao_acc2019} 
\begin{eqnarray}
h(t) &=& \beta_1 \sin \left(\frac{2\pi}{24} t+\phi_1\right) +  \beta_2\sin \left(\frac{4\pi}{24} t+\phi_2\right)
\label{eq:f_seas}
\end{eqnarray}
where $\beta_1, \phi_1, \beta_2, \phi_2$ are model parameters that are estimated in a separate process.
It can be easily verified that $\int_{t=0}^{24} (1+h(t)) \diff t = 1$. Thus, 
the total number of available impressions $N$, corresponding to the case when every 5-min pacing signal equals 1, would be expressed as the following
\bea
N =  4\frac{\sum_{j \in \Omega_a} N_j}{ \sum_{j \in \Omega_a} \left(1+h(j/12) \right)} 
\eea
The above considers the 1:4 sampling rate of internal auction data, and a consistent winning rate $r$ of the DSP in external auction.

\subsubsection{Forecasting}
Once modeled  the event rate distribution as a GMM with parameters $\hat{\bm{\theta}}_e$ obtained in Section \ref{sec:eModel}, we are able to derive the number of impressions and total advertiser cost defined in Eq. (\ref{eq:kpi}). Assuming that event rate $e$ and highest competing price $b^*$ are independently distributed, the joint PDF $p(e, b^*)$ can be written as the product of marginal distribution $p(e)p(b^*)$, where $p(e)$ is the GMM PDF parameterized by $\hat{\bm{\theta}}_e$. The number of impressions can be won under control signal $u$ can be written as follows
\bea 
\label{eq:iu}
n_I(u) &=&NP( f(e,u| \hat{\bm{\theta}}) > b^*)\nonumber \\
&=&N \int_{0}^{\frac{b_{max}-\hat{\theta_0}}{\hat{\theta_1}}} \int_{f^{-1}(b^*, u)}^1 p(e)p(b^*) \diff e \diff b^* \nonumber \\
&=&N  \int_{0}^{\frac{b_{max}-\hat{\theta_0}}{\hat{\theta_1}}} [1-P_e(f^{-1}(b^*, u))] p(b^*)\diff b^* \nonumber \\ 
&=&\frac{N}{N'}\sum_{i =1}^{N'} \left[1-P_e(f^{-1}(b^*_i, u)) \right] \mathbb{I}_{\{ b^*_i \leq \frac{b_{max}-\hat{\theta_0}}{\hat{\theta_1}} \} } 
\eea
where $P(\cdot)$ represents the \textit{Cumulative Distribution Function} (CDF). The last step transforms an integration into a summation, based on \textit{Monte Carlo theory}, knowing that all $b_i^*$ are sampled from its distribution $p(b^*)$ with probability $1/N'$. 
In a similar fashion, the total advertiser spend can be derived as follows
\begin{eqnarray}
\label{eq:cu}
&&c(u) \nonumber \\
&=& N \mbox{E}_{b^{c}, b^*, e}\left[ b^{c} \mathbb{I}_{ \{ f(e,u|\hat{\bm{\theta}}) > b^*\}} \right]\nonumber \\
&=& N\int_{0}^{\infty}\int_{0}^{\frac{b_{max}-\hat{\theta_0}}{\hat{\theta_1}}} \int_ { f^{-1}(b^*, u) }^{1} p(e)p(b^*, b^{c}) b^{c} \diff e \diff b^* \diff b^{c}\nonumber \\
&=&  N \int_{0}^{\infty} \int_{0}^{\frac{b_{max}-\hat{\theta_0}}{\hat{\theta_1}}} \left[1- P_e(f^{-1}(b^*, u)) \right] p(b^*,b^{c})  \diff b^* \diff b^{c} \nonumber \\
&=&  \frac{N}{N'}\sum_{i=1}^{N'}  \left[1-P_e(f^{-1}(b^*_i, u)) \right] b^{c}_i  \mathbb{I}_{\{b^*_i < \frac{b_{max}-\hat{\theta_0}}{\hat{\theta_1}}\}}
\end{eqnarray}

The major advantage of having an analytical form of advertiser spend as a function of control signal in Eq. (\ref{eq:cu}) is to facilitate the calculation of plant gain, which is a significant challenge in the design of a control system for online advertising. It measures the sensitivity of delivery rate (the amount of budget spend in one hour) to the control signal and thus guide the adjustment of control signal to achieve targeted delivery. Based on the definition, which is the first derivative of $c(u)$ with respective to $u$, the plant gain $g(u)$ can be obtained as follows
\bea
&&g(u)  \nonumber\\
&=& \frac{\partial{c(u)}} {\partial{u}}  \nonumber\\
&=& \frac{N}{N'} \sum_{i=1}^{N'}  \left[ - \frac{\partial{P_e(f^{-1}(b^*_i, u))}}{\partial{u}} \right] b^{c}_i  \mathbb{I}_{\{b^*_i < \frac{b_{max}-\hat{\theta_0}}{\hat{\theta_1}}\}} \nonumber\\
&=& \frac{N}{N'}\sum_{i=1}^{N'}  \left[ - p_e(f^{-1}(b^*_i, u)) \frac{\partial{f^{-1}(b^*_i, u)}}{\partial{u}} \right] b^{c}_i  \mathbb{I}_{\{b^*_i < \frac{b_{max}-\hat{\theta_0}}{\hat{\theta_1}}\}} \nonumber\\
&=&\frac{N}{N'}\frac{1}{u^2} \sum_{i=1}^{N'}   p_e(f^{-1}(b^*_i, u))  b^{c}_i  \frac{b^*_i -\theta_0}{\theta_1g} \mathbb{I}_{\{b^*_i < \frac{b_{max}-\hat{\theta_0}}{\hat{\theta_1}}\}} 
\eea
where the last step is by knowing $f(\cdot)$ as a function of control signal $u$ in the bid adjustment model defined in Eq. (\ref{eq:linearModel}). 

To calculate the number of conversions $n_A(u)$ defined in Eq. (\ref{eq:kpi}), we used Monte Carlo method since no close form solution can be found as for $n_I(u)$ and $c(u)$. In this method, we take $N'$ samples from the GMM of event rate $p(e)$, represented by $[e^s_1, \dots, e^s_{N'}]$ and the number events can be expressed as 
\begin{eqnarray}
\label{eq:cu}
n_A(u)&=&  \frac{N}{N'}\sum_{i=1}^{N'} e^s_i \mathbb{I}_{\{ f(e^s_i, u) \geq b^*_i \}}
\end{eqnarray}
The eCPA in Eq. (\ref{eq:kpi}) can be obtained by the following based on the definition 
\bea
v(u) &=& \frac{\sum_{i=1}^{N'}  \left[1-P_e(f^{-1}(b^*_i, u)) \right] b^{c}_i  \mathbb{I}_{\{b^*_i < \frac{b_{max}-\hat{\theta_0}}{\hat{\theta_1}}\}}} {\sum_{i=1}^{N'} e^s_i \mathbb{I}_{\{ f(e^s_i, u) \geq b^*_i \}}}
\eea


\subsection{Sampling}
The internal auction data is a huge amount of data. Is it necessary to use all the data to make forecast? If not, how many samples we need for a desired confidence level? This subsection is to answer these questions.

We define a random variable $x$ which is Bernoulli distributed
\begin{eqnarray}
x=\left\{
            \begin{array}{ll}
              1,~~P(b \geq b^*)\\
              0, ~~1- P(b \geq b^*).
            \end{array}
          \right.
\end{eqnarray}
Basically, for a given control signal $u$, if the bid price for impression $i$ is greater than the highest competing bid, then a realization of random variable $x_i = 1$, otherwise $x_i = 0$. So, for any value $u$, an estimator of $P$ (we us $P$ as an abbreviation for $P(b \geq b^*)$ in the remaining of this subsection, but knowing that it is a variable depending on the control signal $u$) would be the average values of $x_i, \forall i \in \{1, \dots, N\}$, i.e., 
\bea
\hat{P} = M = 1/N \sum_{i= 1}^{N } x_i 
\eea
According to \textit{Central Limit Theory} (CLT), when independent random variables are added, their properly normalized sum tends to a normal distribution. Thus, we have 
\bea
\label{eq:M}
M \sim \mathcal{N} (P, \frac{P(1-P)}{N})
\eea
In our case, $P$ is the parameter to be estimated for every possible values of $u$. A confidence interval for the parameter $P$ with confidence level $\epsilon$, determined by $\gamma$,  is an interval with the following property 
\bea
P(\|P-M \| \geq \epsilon ) \leq \gamma
\eea
Knowing the distribution of $M$ in Eq.(\ref{eq:M}), the left hand of the above can be written as follows
\bea
P(\|P-M \| \geq \epsilon ) &=& 2-2\Phi \left(\frac{\epsilon \sqrt{N}}{\sqrt{P(1-P)}} \right)  \nonumber \\
&\leq&2-2\Phi \left(2\epsilon \sqrt{N}\right) 
\eea
where $\Phi(\cdot)$ denotes the CDF of Gaussian distribution and
the last step is by knowing that $P(1-P)$ achieves its maximum at $P= 1/2$ for $P \in [0,1]$. To have the following confidence level satisfied
\bea
2-2\Phi \left(2\epsilon \sqrt{N} \right)  \leq \gamma
\eea
we can derive that 
\bea \label{eq:N}
N \geq \left(\frac{1}{2\epsilon} \Phi^{-1}\left((1 - \gamma/2) \right) \right)^2
\eea
According to the above inequality, when choosing $\gamma = 0.95$ and $\epsilon = 0.01$, meaning that with probability $95\%$, the estimated $M$ is within a very small vicinity of the ground true values of $P$, i.e., $M \in [P-0.01, P+0.01]$, we need to take 10,000 samples from each line. Eq. (\ref{eq:N}) provides a guidance on sampling a sufficient number of samples from the raw internal auction data set based on desired confidence level of the impression vs. control curve, i.e., $n_I(u)$.

\section{Experimental Results} \label{sec:results}
In this section, we demonstrate how we applied the forecasting approach to generate the key performance metrics using one day historical data for eCAP lines. And we further validated our forecasting results.
\subsection{Forecast Results }\label{sec:forecastR} 

The actual control signal in production is in the range of $[10^{-6}, 50]$, but the active range can be quite different from line to line. So for our forecast to be most informative, we first identify the active range of control signal for a line $[0, u_{max}]$, where 
\[u_{max} = \max{ \{ \frac{ b^*_i}{ge_i}, \forall b^*_i \leq \frac{b_{max} -\hat{\theta}_0}{\hat{\theta}_1}\}}
\] corresponds to the control signal value above which no more impressions can be won.

\begin{figure}
\centering
\includegraphics[width=.7 \columnwidth,height=!]{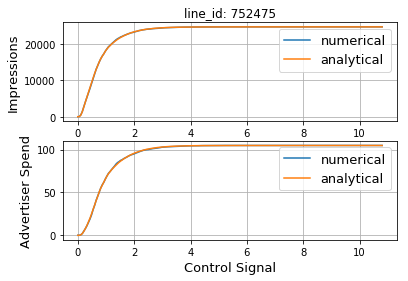}
\caption{Number of impressions and total advertiser spend as functions of control signal}
\label{fig:uiuc}
\end{figure}

The control response curve defined in Eq. (\ref{eq:iu}) and (\ref{eq:cu}) is shown in Fig \ref{fig:uiuc}. As can be seen from the equations, the response curves are monotonically increasing function of control signal. We also compared the curves calculated in Eq. (\ref{eq:iu}) and (\ref{eq:cu}) with numerical method, where a set of event rates are resampled from the GMM and number impressions and spend are counted numerically for every control signal. As shown in the figure, curves by the two methods are identical. 

\begin{figure}
\centering
\includegraphics[width=.7\columnwidth,height=!]{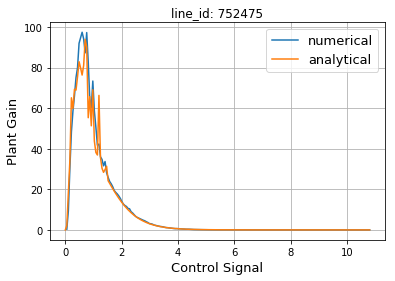}
\caption{Plant gain as a function of control signal}
\label{fig:plantgain}
\end{figure}

The plant gain is plotted out in Fig. \ref{fig:plantgain}, along with the plant gain calculated numerically which is a discrete representation of the function and can not be directly consumed by the controller and only serves as a reference for the analytical plant gain. It can be seen that the two curves, analytical and numerical, are almost overlap with each other. For control  signal within $[0.1, 1.7]$, the plant gain is relatively high, corresponding to the steep changes of both impressions and advertiser spend in Fig. \ref{fig:uiuc}. Such information is used in the feedback control to determine how aggressive the control signal needs to change. 

\begin{figure}
\centering
\includegraphics[width=.7\columnwidth,height=!]{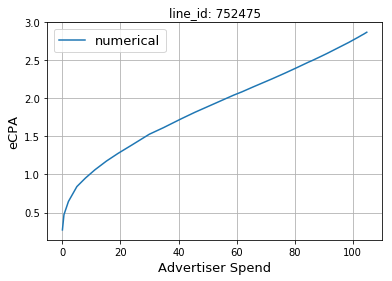}
\caption{Relationship between advertise spend and eCPA}
\label{fig:spend_ecpa}
\end{figure}

To advertisers, the plot in Fig. \ref{fig:spend_ecpa} gives insight on how much incremental budget they can spend with the DSP without eCPA exceeding the goal amount. According to the Optimization Mission Control \footnote{\path{yo/opmt-missioncontrol}}, there are around 39\% CPA lines who has a delivery ratio $>0.8$ and a performance ratio way below $1$. Those lines can benefit from the results in  Fig. \ref{fig:spend_ecpa} and suggestions can be made to the advertisers about much more budget we are able to spend safely. 

\subsection{Validation}
\begin{figure}
\centering
\includegraphics[width=.7\columnwidth,height=!]{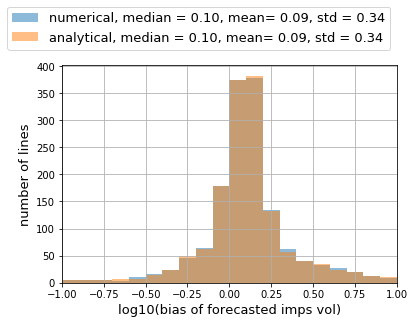}
\caption{Histogram of forecasting bias in logarithm }
\label{fig:bias}
\end{figure}

To validate the forecasting results, we compared the forecasted number of impressions corresponding to the actual control values of day $k+1$, denoted by $u_{k+1}$, with the actual delivered impressions $I'_{k+1}$. Note that, since the actual delivery corresponds to a pacing signal vector which may not be a all 1 vector, so we first normalize the actual delivery using the same method in Section \ref{sec:normalization} and get a normalized actual delivery $I_{k+1}$ .We further define the forecasting bias, denoted by $\rho$, as the ratio between the two, i.e., $\rho = \frac{I(u_{k+1})}{I_{k+1}}$. The validation results is plotted in Fig. \ref{fig:bias}, where the bias $\rho$ is in logarithm to make over-forecast and under-forecast equally visible. It can be seen that the forecast is almost unbiased, and with probability 90\%, the forecast is within $[0.339, 4.459]$ times the actual delivery. And with probability 50\%, the forecast is within $[0.726, 2.08]$ times the actual delivery, which is a very good performance considering that such forecasting for eCPA lines is first available within Verizon Meidia DSP and the existing system can tens of times under-forecast or over-forecast for eCPM lines. 

\section{Implementation}\label{sec:imp}
To implement the forecasting algorithm in production, we utilize Apache oozie workflow scheduler system to manage our job pipeline that can generate forecasting result for each line from $400$GB raw data everyday. The workflow has three components: read and preprocess internal auction/controller data from kite and control report, run the forecasting algorithm and generate forecasting metrics, and finally validate last day’s forecasting result. Currently, the data pipeline is running and we are working refining the data-flow and enabling monitoring and visualization.

\section{Conclusion and Future Work}\label{sec:conclusion}
In this paper, we proposed a novel approach for mid-flight forecasting for CPA lines by taking bidding strategy into account. We forecasted the number of impressions, total advertiser spend as analytical functions of the control signal, based on which we derived the plant gain, a key information metric for the design of feedback control for online advertising. The relationship between advertiser spend and eCPA performance is also investigated which provides advertisers actionable insights. The forecast results were validated against the actual delivery. In the future, we will generalize the method to other goal type lines. We will also expend our forecasting method to pre-flight which is a more challenging case due to the lack of line specific historical data. 

\bibliographystyle{ACM-Reference-Format}	
\bibliography{bibli}		

\end{document}